\documentclass{article} % For LaTeX2e
\usepackage{iclr2016_conference,times}
\usepackage{hyperref}
\usepackage{url}

\usepackage{amsmath}
\usepackage{verbatim}
\usepackage{amssymb}
\usepackage{amsthm}
\usepackage{color}
\usepackage{graphicx}
\usepackage{subfigure}

\newcommand{\ignore}[1]{}

\title{Learning to Represent Words in Context\\ with Multilingual Supervision}

\author{Kazuya Kawakami, Chris Dyer\\
Department of Computer Science\\
Carnegie Mellon University\\
Pittsburgh, PA 15213, USA \\
\texttt{\{kkawakam, cdyer\}@cs.cmu.edu} \\
}

\begin{document}

\maketitle

\begin{abstract}
We present a neural network architecture based on bidirectional LSTMs to compute representations of words in the sentential contexts. These context-sensitive word representations are suitable for, e.g., distinguishing different word senses and other context-modulated variations in meaning. To learn the parameters of our model, we use cross-lingual supervision, hypothesizing that a good representation of a word in context will be one that is sufficient for selecting the correct translation into a second language. We evaluate the quality of our representations as features in three downstream tasks: prediction of semantic supersenses (which assign nouns and verbs into a few dozen semantic classes), low resource machine translation, and a lexical substitution task, and obtain state-of-the-art results on all of these.
\end{abstract}

\section{Introduction}
Distributed representations of words, which represent each word as a vector in a low-dimensional space, can be learned from unannotated text corpora using a variety of techniques~\citep{mikolov2013efficient,pennington2014glove,landauer:1997}. The value of such representations owes to their ability to capture intuitive notions of syntactic and semantic similarity as geometric locality. Despite their empirically proven value as a source of features in many downstream applications~\citep{turian2010word}, the ``one word type, one vector'' assumption made by most word representation models is problematic because words may have multiple meanings.

Two standard solutions to this problem exist. The first to treat each word as a collection of discrete, mutually exclusive senses which are individually represented as vectors~\citep{zhang:2014,neelakantan2014efficient,wu:2015,huang:2012,jauhar:2015}. However, identifying the appropriate sense granularity in such models is difficult in practice and in theory~\citep{kilgarriff1997don,erk:2013}. The second solution, which is the basis of this work, eschews sense inventories (whether latent or explicit) and says that lexical meaning is a function of word and its context~\citep{erk:2008,kintsch:2001,mitchell:2008}. While previous work has hinted at the promise of this solution, only a small number of hand-crafted word--context composition functions have been considered thus far in the literature on semantic representation learning. This is surprising given the success of learning composition functions for computing phrase and sentence representations~\citep{socher:2011,kalchbrenner2014convolutional}.

There are two central challenges faced by learning to represent words in context. The first is to identifying a suitable function class for the composition function. Such a function must be able to account for the fact that a single word type may have both several completely unreleated meanings as well as a several more or less distinct but still related meanings \citep{cruse2000meaning}. For an example of the former, the word \emph{plant} may refer, depending on context, to a factory or to a living organism that photosynthesizes. For an example of the latter, the word \emph{bank} may refer to a financial institution or the building housing a financial institution. Since bidirectional RNN-LSTMs have been shown to be able to learn both compositional~\citep{bahdanau2014neural} as well as more arbitrary relationships~\citep{ling2015finding}, we use these as our composition function class (\S\ref{sec:model}).

The second challenge is to identify an appropriate supervisory signal that will be used to fit the parameters of the function. Our motivating hypothesis---which follows a long line of work in using parallel data as a source of information about semantics~\citep{bannard:2005,resnik:1999,diab:2003,faruqui2014improving,hermann:2014}---is that a good representation of a word in context will be one that predicts how that word (in its sentential context) translates into a second language~(\S\ref{sec:background}). We show that word-in-context representations can be learned efficiently from pairs of words-in-context and single word translations into a second language which are extracted from parallel corpora using a word alignment model.

To evaluate our proposed model and training criterion, we evaluate our learned representations as features in three tasks: supersense tagging, low-resource machine translation (i.e., translation where limited parallel data is available), and a lexical substitution task. Success in each of these requires models that can effectively capturing the meaning of a word in context, and in each, we show our model obtains state-of-the-art performance (\S\ref{sec:experiments}). Additionally, the feedforward neural net model we use as a baseline for supersense tagging outperforms existing baselines even without our new word-in-context model.

\section{Model}\label{sec:model}
Our model for contextual words is a bidirectional sequence model based on recurrent neural networks \citep[\emph{inter alia}]{chan:2015,bahdanau2014neural,bahdanau:2015b}. Intuitively, this model allows us to condition on arbitrarily long dependencies while having an implicit bias toward more local contexts.

Let $\boldsymbol{w}=(w_{1}, w_2, \ldots, w_{n})$ be the words in a sentence with length $n$. We also project all words into a fixed $d$-dimensional vectors $\boldsymbol{x} = (\mathbf{x}_{1}, \mathbf{x}_2, \ldots, \mathbf{x}_{n})$, using a (one-word-per-type) word lookup table.

The model encodes each token of the sentence from left to right according to the standard Long-short term memory recurrences:
\begin{align*}
\mathbf{i}_{t} &= \sigma(\mathbf{W}_{xi}\mathbf{x}_{t} + \mathbf{W}_{hi}\mathbf{h}_{t-1} + \mathbf{W}_{ci}\mathbf{c}_{t-1} + \mathbf{b}_{i})\\
\mathbf{f}_{t} &= \sigma(\mathbf{W}_{xf}\mathbf{x}_{t} + \mathbf{W}_{hf}\mathbf{h}_{t-1}+\mathbf{W}_{cf}\mathbf{c}_{t-1} + \mathbf{b}_{f})\\
\mathbf{c}_{t} &= \mathbf{f}_{t}\odot\mathbf{c}_{t-1} + \mathbf{i}_{t}\odot \tanh(\mathbf{W}_{xc}\mathbf{x}_{t} + \mathbf{W}_{hc}\mathbf{h}_{t-1} + \mathbf{b}_{c})\\
\mathbf{o}_{t} &= \sigma(\mathbf{W}_{xo}\mathbf{x}_{t} + \mathbf{W}_{ho}\mathbf{h}_{t-1} + \mathbf{W}_{co}\mathbf{c}_{t} + \mathbf{b}_{o})\\
\mathbf{h}_{t} &= \mathbf{o}_{t}\odot\tanh(\mathbf{c}_{t})
\end{align*}
This yields a representation $\overrightarrow{\mathbf{h}_{t}}$ for each position in the sentence $t$ which can be interpreted as the representation of word with its left context $w_{1}, w_2, \ldots, w_t$. The same process is repeated from right to left, yielding a vector $\overleftarrow{\mathbf{h}_{t}}$. The concatenation of these two vectors 
\begin{align*}
\mathbf{h}_{t} = [\overrightarrow{\mathbf{h}_{t}} ; \overleftarrow{\mathbf{h}_{t}}],
\end{align*}
is our word-in-context representation.

\subsection{Model intuition}
\begin{figure}[ht]
\begin{center}
\includegraphics[width=0.75\linewidth]{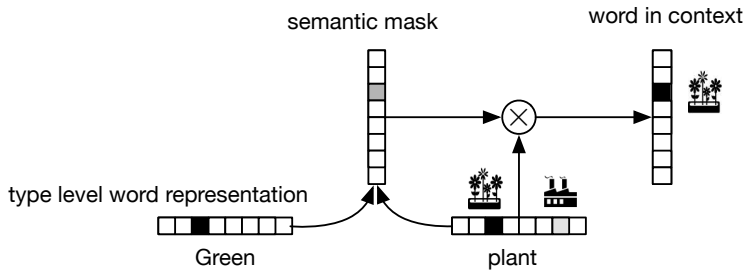}
\end{center}
\caption{Description of the operation to modulate sense from ambiguous type level vector.}
\label{fig:wordsense}
\end{figure}

Type-level word embeddings must necessarily represent information about multiple senses in a single vector, and our task is to obtain. To obtain a representation of a word in its context, we want to apply functions which mask or scale some dimensions of the vector according to its context. Thus, functions which apply same scaling function even the word and context are different, average and multi-layer perceptron for example, may not be suitable. The input gate in Long-short term memory is considered to be a suitable scaling function which take target word and its context ($\mathbf{x}_{t}, \mathbf{h}_{t}, \mathbf{c}_{t-1}$). Figure \ref{fig:wordsense} show a simplified version of operation to modulate one sense (vegetable plant) from ambiguous type level vector with semantic mask which is conditioned on word and context.

\section{Meaning and Translation}\label{sec:background}
We now require a training objective that provides supervision for learning the parameters of this model.
The question we want to answer is: what is a suitable proxy (or ``grounding'') for the meanings of words in context that we can use to construct \emph{token-level} (rather than type-level) word representations?

To illustrate the problem we wish to solve, consider the meaning of the token \emph{bank} in the following sentences:
\begin{itemize}
\item I went to the \textbf{bank} to deposit my paycheck.
\item I went to the river \textbf{bank} to eat some lunch.
 \end{itemize}

One very productive strategy for learning semantic word embeddings is to rely on the distributional hypothesis \citep{harris-54}, according to which semantically similar items occur in similar contexts. The distributional hypothesis is, furthermore, practically appealing since it enables semantics to be learned from large, unannotated text corpora.
 
Despite the empirical success of the distributional hypothesis at obtaining representations of word \emph{types}, creating a representation of word \emph{tokens} in terms of context is conceptually unappealing since both the item being embedding and its context potentially share material. One possible solution would be an autoencoding objective, or one might also distinguish between ``narrow'' and ``wide'' context (i.e., one that determines the item being embedding and one that provides supervision).\footnote{For example, \cite{mikolov:2013} showed that short multiword expressions could be embedding by using the ``wider'' context that they occur in.}

However, we instead advocate using an alternative proxy for meaning: how words \emph{translate}. Consider the two examples from above as they might be translated into French.
\begin{itemize}
\item Je suis all\'e \`a la \textbf{banque} pour d\'eposer mon ch\`eque de paie.
 \item Je suis all\'e sur la \textbf{rive} pour le d\'ejeuner.
 \end{itemize}
The homonymous (i.e., having two completely unrelated senses) word \emph{bank} has been translated into two different words \emph{banque} and \emph{rive} in French.

Finally, while not quite so copious as monolingual corpora, parallel data exist in convenient electronic form in abundance, and this provides a rich resource for learning about the semantics of natural language.

\subsection{Objective \& Parameter Learning}\label{sec:learning}
To operationalize our hypothesis that translation provides a good supervisory signal for learning semantic representations, we learn the parameters of source language word type embeddings and the composition function (i.e., the parameters of the bidirectional LSTMs) by using the computed representation to compute the lexical translation probability of a word in context. That is, we use the computed token embedding  to define a probability estimate that a source language word $e_t$ in context $\boldsymbol{c}=(e_1,\ldots,e_{t-1},e_{t+1}\ldots,e_n)$ translates into a second language as $f$ in vocabulary $\mathcal{F}$. i.e., $p(f \mid e_{t},\boldsymbol{c})$.

This is done by performing a softmax over the target vocabulary with the representation of the word $\mathbf{h}_t$, as defined in the previous section. That is, we compute
\begin{align*}
\mathbf{u} &= \mathbf{Rh}_t + \mathbf{b}'\\
p(f \mid e_t, \boldsymbol{c}) &= \frac{\exp(u_{f})}{\sum_{f' \in \mathcal{F}}\exp(u_{f'})},
\end{align*}
where parameters $\mathbf{R}$ and $\mathbf{b}'$ define the projection of the source word with context representation $\mathbf{h}_t$ onto the target vocabulary $\mathcal{F}$.

To obtain pairs of words in context and their lexical translations into a second language, we use unsupervised word alignment techniques~\citep{dyer2013simple}, to obtain high precision word alignments from a parallel corpus. While modeling alignments as latent variables, or using a soft attention mechanism would be a reasonable alternative, word alignment is fast and the proposed training objective to be easily scaled to large corpora.

\begin{figure}[ht]
\begin{center}
\includegraphics[width=0.8\linewidth]{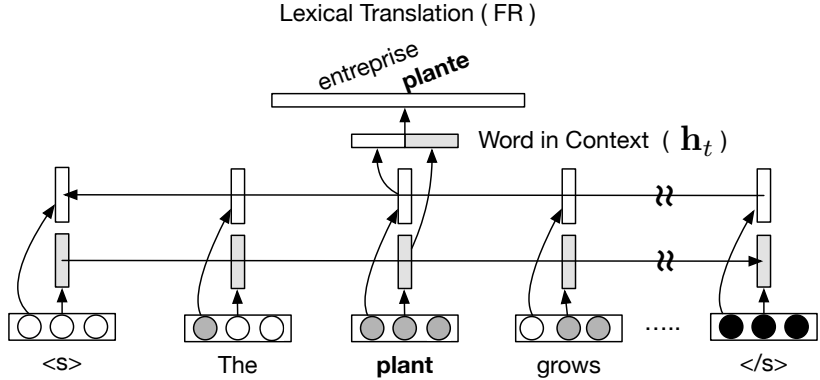}
\end{center}
\caption{Description of cross lingual pre-training model.}
\label{fig:model}
\end{figure}

Figure~\ref{fig:model} illustrates the pre-training architecture.

\subsection{Parameter Learning}
The model parameters $\mathbf{W}$ and $\mathbf{b}$ as well as the word projection parameters $V_{e}$ are first pre-trained with the objective function:
\begin{align*}
\mathcal{L} = - \sum_{(\boldsymbol{f}, \boldsymbol{e})} \log p(f \mid e, \boldsymbol{c})
\end{align*}
That is, we wish to find the parameters that maximize the lexical translation log probability over the whole parallel corpus of lexical translations ($f$) of a source word ($e$) in context ($\boldsymbol{c}$). 

When we want to transfer the model to another supervised task to predict label $s \in \mathcal{S}$ for a word $e$ in context $\boldsymbol{c}$, the final values of the $\mathbf{W}$ and $\mathbf{b}$ parameters are transferred and formulate a similar model to predict label $s$. Using the transformation matrix $\mathbf{S} \in \mathbb{R}^{|\mathcal{S}| \times d_{h}}$ and the biases $\mathbf{b}'' \in \mathbb{R}^{|\mathcal{S}|}$, we may define the label probability as
\begin{align*}
\mathbf{u}' &= \mathbf{Sh}_{t} + \mathbf{b}''\\
p(s \mid e_t, \boldsymbol{c}) &= \frac{\exp(u'_{s})}{\sum_{s^{'} \in \mathcal{S}}\exp(u'_{s'})}.
\end{align*}

 the model is training by maximizing the log likelihood of the observed label in the task. 
\begin{equation}
\mathcal{L}' = - \sum_{(\boldsymbol{s}, \boldsymbol{e})} \log p(s \mid e, \boldsymbol{c})
\end{equation}

\section{Experiments}\label{sec:experiments}
We now turn to a series of experiments to show the value of learning representations of words in context according to the objective above. Our paradigm will be to pre-train using the objective above the parameters of a word-in-context model, and then use these (without further fine tuning) in downstream tasks: prediction of semantic supersenses (which assign nouns and verbs into a few dozen semantic classes), low resource machine translation, and a lexical substitution task.

\subsection{Model configuration and pre-training}
To pre-train our model, we extracted words ($e$) in contexts and translations ($f$) from the Europerl parallel corpus~\citep{koehn2005europarl}. We conducted experiments with the following four languages: French (FR), German (DE), Czech (CS), Finnish (FI) which are quite typologically diverse. Table\ref{tb:parallel} shows the numbers of parallel sentences and the numbers of words. For each language pairs, we used 2000 sentences for development and the rest were used for training. 

After normal tokenization, we obtained alignments with fast-align tool~\citep{dyer2013simple}. Since we are modeling single word translations and want high-quality training instances, we run the alignment model in both directions and obtained symmetric alignments by taking intersection between forwards and backward alignments. To control the size of vocabulary, we took 30,000 most common words. For target languages, we removed 10 most common words. The words not in the vocabularies are replaced with $\langle$unk$\rangle$ token. We used sentences which have more than 10 words in a sentence.

We used 300 dimension embeddings for source language, and bi-directional LSTMs have 300 hidden units. The trained parameters are source embedding, weights and bias in the model.

We randomly initialized source word embeddings sampled from uniform distribution from $-0.08$ to $0.08$. All recurrent materices with orthogonal initialization~\citep{andrew2013exact}, and non-recurrent weights are initialized from scaled uniform distribution~\citep{glorot2010understanding}. Mini-batches of size 128 are used. We used Adam algorithm for optimization~\citep{kingma2014adam}. We trained models with early-stopping. The perplexities on development data for English to French, German, Czech and Finnish are 3.80, 6.49, 6.30, 19.25 respectively. 

\begin{table}[t]
\caption{Summary of parallel data.}
\label{tb:parallel}
\begin{center}
\begin{tabular}{lcrrrrr}
\multicolumn{1}{c}{\bf Dataset} &\multicolumn{1}{c}{\bf Data Source} & \multicolumn{2}{c}{\bf Vocabulary} &\multicolumn{2}{c}{\bf Token} &\multicolumn{1}{c}{\bf Sentence} \\\hline
 & &\multicolumn{1}{c}{\bf Source} &\multicolumn{1}{c}{\bf Target} & \multicolumn{1}{c}{\bf Source} &\multicolumn{1}{c}{\bf Target} & \\ \hline
EN-FR & europarl-v7 & 93,393 & 139,934 & 50,586,497 & 52,900,470 & 1,835,733 \\
EN-DE& europarl-v7  & 93,033 & 323,367 & 48,625,466 & 46,484,368 & 1,763,744 \\
EN-CS & europarl-v7 & 51,833 & 164,770 & 16,150,983 & 13,785,699 & 594,158 \\
EN-FI & europarl-v7   & 91568 & 630184 & 48,584,379 & 34,819,783 & 1,779,397 \\
EN-MG & CMU          &  57,668 & 76,469 &  1,592,66 & 2,023,336 & 80,306 \\
EN-UR & NIST MT08 &  43,524 & 44,566 &  1,055,030 & 1,180,031 & 161,173 \\
\end{tabular}
\end{center}
\end{table}

\subsection{Supersense Tagging}\label{sec:supersense}
Supersenses can be thought of a generalization of words senses into a universal inventory of semantic types. That is, as the number of word senses tend to be too numerous for existing models to generalize properly with the small amounts of data available, supersenses address this problem by clustering all senses into a tractable set of tags. Table \ref{tb:sstags} show examples of supersense tags and its definition. As such, these are generally used in semantically oriented downstream tasks such as co-reference resolution~\citep{DBLP:journals/corr/OConnorH13} and question answering~\citep{pasca2001informative}.

Following previous work, we trained our supersense tagger for nouns and verbs on the Semcor dataset. The Semcor datasets consists of three parts, \textsf{brown1}, \textsf{brown2}, and \textsf{brownv}. We mixed these three parts and trained supersense tagger on randomly split 4/5 of data and the rest were used as a development set. We evaluated our model on the held-out SensEval-3 all-words task~\citep{mihalcea2004sel}, as done in previous work on supersense tagging~\citep{ciaramita2006broad,yuret2010noisy}. Since some tokens are annotated with two labels in ambiguous cases, we followed the heuristics of only using the first sense in the data as the correct synset/supersense~\citep{ciaramita2006broad}. To extract supersenses from the Semcor data, we used WordNet version~2.0 synsets.

To avoid the computational overhead of reading extremely wide contexts, we used sliding window to delimit the range of contexts as in~\citep{collobert2011natural}, that is, each token $w_t$ is embedded using a context window of words $w_{t-n/2}, \ldots, w_{t}, \ldots, w_{t+n/2}$. The window size $n$ was fixed to 20.

We use the pre-trained parameters and we put a new task specific softmax leyer on top of the hidden units (Fig.\ref{fig:model}). We updated all parameters including pre-trained parameters. The weights in the softmax layer were initialized from the scaled uniform distribution ~\citep{glorot2010understanding}. Mini-batches of size 128 were used with the Adam update rule~\citep{kingma2014adam}. 

Since this task has not previously been studied using neural networks, we also report several novel baselines: (1) multi-layer perceptron model which uses a concatenation of a source word type vector and the average of all word type vectors in its context; (2) a forward-only LSTM model; and (3) a bi-directional LSTM with random initialization (rather than cross-lingual pretraining). For fair comparison in terms of the size of word in context representation, we double the hidden unit size of the forward LSTM model.

\begin{table}[thb]
  \centering
  \small
  \renewcommand{\arraystretch}{1.1}
  \begin{tabular}{lp{1.7in}|lp{1.7in}}
    Supersense & Nouns denoting & Supersense & Verbs denoting \\ \hline
    act & acts or actions & change & size, temperature change\\
    artifact & man-made objects & communication & telling, asking, ordering, singing \\
    feeling & feelings and emotions & possession & buying, selling, owning \\
    group & groupings of people or objects & plant & plants \\
    location & spatial position & social & political and social activities
  \end{tabular}
  \label{tb:sstags}
  \ignore{\caption{26 Noun Supersenses and 15 Verb Supersenses}}
  \caption{Examples of Noun and Verb Supersenses}
\end{table}

\subsection{Lexical Translation in Low Resource Language}
We investigate the benefit to transfer cross lingually pre-trained word-in-context representation to translation in low-resource language. Since low-resource languages do not have enough data to adequate estimate translation probabilities, we hope that we can learn more effective mappings with pre-trained word-in-context embeddings ~\citep{chahuneau2013translating}.

We trained lexical translation model, which predict translation of aligned English sentence, for low resource languages, Malagasy and Urdu on top of the pre-trained word in context model. Table \ref{tb:parallel} shows the numbers of parallel sentences and the number of words. We used a dataset used in~\citep{dou2014beyond} for Malagasy and the Urdu data we used is a part of NIST MT evaluation in 2008-2012\footnote{https://catalog.ldc.upenn.edu/LDC2010T21}. We used 2000 sentences for development and hold-out test set. We filtered out sentences which have less than 3 words for pre-training and words occur less then 1 time are replaced with $\langle$unk$\rangle$ token.

We trained our baseline system with cdec~\citep{dyer2010cdec} and obtained synchronous context-free grammars rules to translate sentences. We added features, translation probability and log translation probability from our translation model and optimized the parameters of a machine translation system with MIRA, Margin-Infused Relaxed Algorithm~\citep{crammer2003ultraconservative}.

\subsection{Lexical Substitution}
Lexical substitution is the problem of identifying meaning-preserving substitutes for a target word given a sentential context. The task was introduced in SemEval-2007~\citep{mccarthy2007semeval} involves both finding the synonyms and disambiguating the context. As such, it is an ideal test case for our representations.

Models are evaluated on their ability to predict the substitutes in the gold standard of the LS-SE test-set. We evaluated our model on best and best-mode task which evaluate the quality of the best predictions. The original task allow to make multiple predictions but we only predict only one substitution following~\citep{melamud2015simple}. This task is challenging, since it requires to find the best substitutes from entire word vocabulary.

The way to make prediction is the following. Given a target word and it's context, we infer word in context representation of all possible substitutions. Then take one of the most similar words which have highest cosine similarity with target word in context vector as prediction.

For our experiments, we used a simple word alignment base candidate generation to reduce inference time. For a target word in English, we collect all possible French translations from word alignment and took English words 90\% most frequently aligned to the French words as candidates. We used same candidates for all our experiments including baseline for fair comparison.

\section{Result}
\begin{table}[t]
\caption{Summary of results for supersense tagging.}
\label{tb:result-supersense}
\begin{center}
\begin{tabular}{l|ccc|ccc}
 &\multicolumn{3}{c}{\bf Semcor} &  \multicolumn{3}{c}{\bf Senseval3}\\\hline
\bf{Method} & \bf{Precision} & \bf{Recall} & \bf{F1} & \bf{Precision} & \bf{Recall} & \bf{F1}\\\hline
Random         & 38.2 & 43.0 & 40.4 & 35.8 & 42.1 & 38.7  \\
Baseline         & 63.9 & 69.3 & 66.5 & 60.1 & 68.7 & 64.1  \\
HMM              & 76.7 & 70.5 & 77.7 & 67.6 & 73.7 & 70.5  \\
CRF               & 80.3 & 80.2 & 80.2 & -       & -       & -        \\\hline
MLP               & 82.0 & 81.9 & 81.6 & 83.6 & 79.7 & 81.2  \\
LSTM             & 82.1 & 82.6 & 82.1 & 83.8 & 81.9 & 82.5  \\
bi-LSTM         & 83.5 & 84.2 & 83.6 & 84.6 & \bf{82.9} & 83.3  \\\hline 
bi-LSTM (FR) & 84.8 & 85.0 & 84.7 & 85.8 & 82.8 & 84.0 \\
bi-LSTM (DE) & 85.2 & 85.2 & 85.0 & 86.2 & 82.7  & 84.1 \\
bi-LSTM (CS) & 84.9 & 85.0 & 84.7 & 85.9 & 82.8  & 84.1 \\
bi-LSTM (FI)   & 85.0 & 85.1 & 84.8 & 85.9 & 82.4  & 83.9 \\\hline
bi-LSTM (average) & \bf{85.0} & \bf{85.1}& \bf{84.8} & \bf{86.0} & 82.7 & \bf{84.0}\\
\end{tabular}
\end{center}
\end{table}

\begin{table}[t]
\caption{Summary of results for Translation in low resource Languages.}
\label{tb:result-translation}
\begin{center}
\begin{tabular}{l|cc|cc}
 &\multicolumn{2}{c}{\bf MG} &  \multicolumn{2}{c}{\bf UR}\\\hline
Method         & \bf{Perplexity} $\downarrow$ & \bf BLEU $\uparrow$ 
			   & \bf{Perplexity} $\downarrow$ & \bf BLEU $\uparrow$ \\\hline
bi-LSTM (random init) & 16.17   & 21.7     & 30.87        & 21.2\\\hline
bi-LSTM (FR)             & 12.80    & 21.9     & 26.84        & 21.7\\
bi-LSTM (DE)             & 12.97   & 21.9      & 26.38        & 21.4\\
bi-LSTM (CS)             & 13.05   & 22.0      & 26.21        & 21.4\\
bi-LSTM (FI)               & 13.07   & 22.0      & 25.96        & 21.5\\\hline
bi-LSTM (average)& \bf{12.97} & \bf{22.0} & \bf{26.34} & \bf{21.5} 
\end{tabular}
\end{center}
\end{table}

\begin{table}[t]
\caption{Summary of results for Lexical Substitution.}
\label{tb:result-lexsub}
\begin{center}
\begin{tabular}{l|cc}
\bf{Method} & \bf{best} $\uparrow$ & \bf{best mode} $\uparrow$ \\\hline
Base                        & 7.81  & 13.41\\
Mult                         & 6.64  & 10.89\\
BalMult                    & 8.09  & 13.41\\
Add                          & 7.37  & 12.11\\
BalAdd                     & 8.14  & 13.41\\\hline
Skipgram (baseline)& 7.77  & 13.16 \\
bi-LSTM (FR)          & 9.54  & 15.79\\
bi-LSTM (DE)          & 10.63& 18.09\\
bi-LSTM (CS)          & 9.74  & 16.04\\
bi-LSTM (FI)            & 8.51  & 12.99\\\hline
bi-LSTM (average)  & \bf{9.60} & \bf{15.73}\\
\end{tabular}
\end{center}
\end{table}

\begin{table}[t]
\caption{Disambiguation with multilingual supervision.}
\label{tb:example-lexsub}
\begin{center}
\begin{tabular}{l|l}
\multicolumn{1}{c}{\bf Sentence} & \multicolumn{1}{c}{\bf Translation candidate for \textit{\textbf{plant}}}\\\hline
They built a large \textit{\textbf{plant}} to manufacture automobiles. & usine, installation, plante, centrale\\\hline
Let's \textit{\textbf{plant}} flowers in the garden. & plantes, planter, v\'{e}g\'{e}tal, v\'{e}g\'{e}tale, cultiver\\
\end{tabular}
\end{center}
\end{table}

\ignore{\begin{figure}[t]
	\begin{tabular}{cc}
	\begin{minipage}{0.5\hsize}
	\begin{center}
		\includegraphics[width=0.8\linewidth]{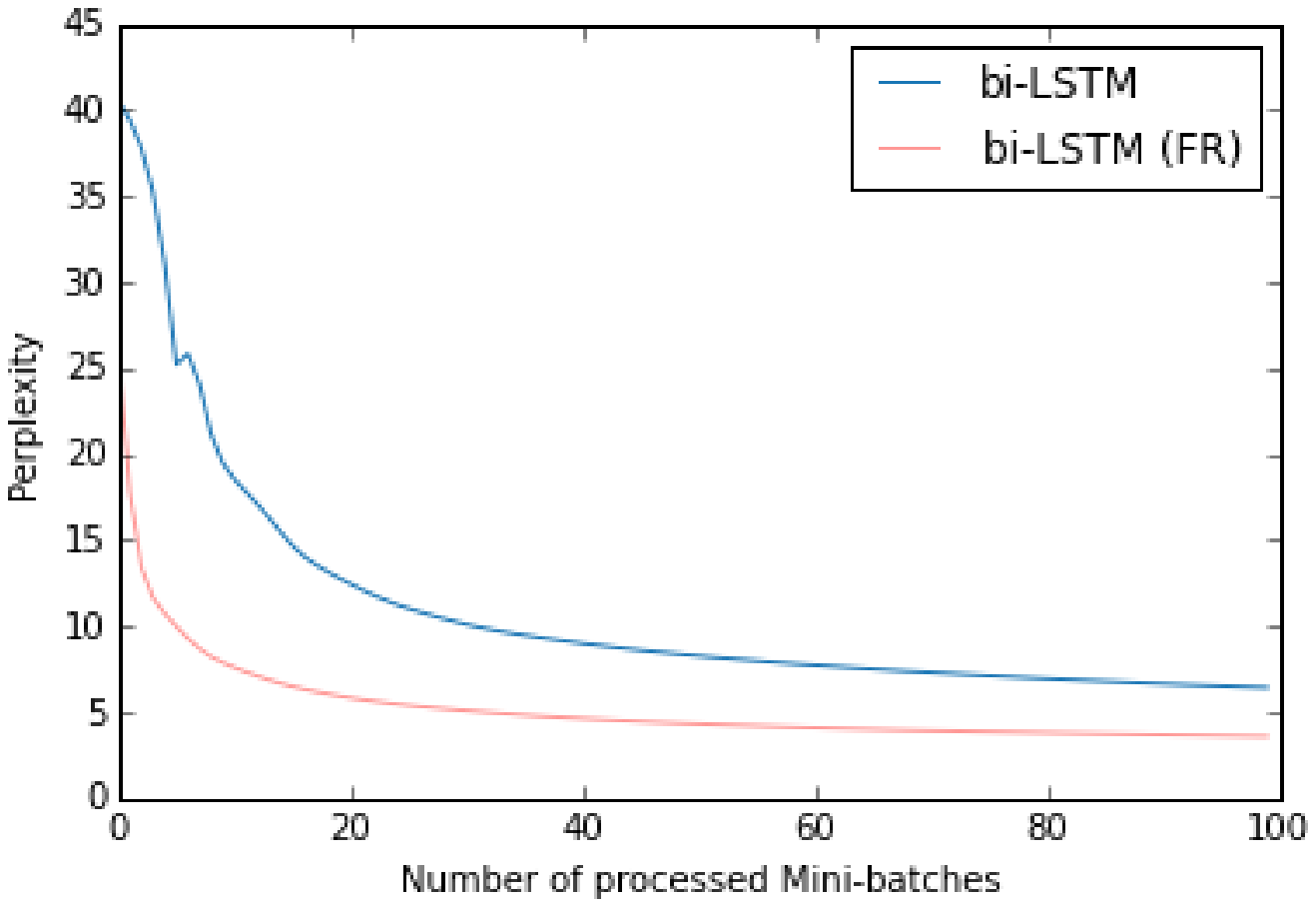}
	\end{center}
	\end{minipage}
	\begin{minipage}{0.5\hsize}
	\begin{center}
		\includegraphics[width=0.8\linewidth]{supersense.eps}
	\end{center}
	\end{minipage}
	\label{fig:fall}
	\caption{Effect of pre-training for low resource machine translation.}
	\end{tabular}
\end{figure}}

\begin{figure}[t]
    \subfigure{
        \includegraphics[clip, width=0.5\columnwidth]{supersense.eps}
    }
    \subfigure{
        \includegraphics[clip, width=0.5\columnwidth]{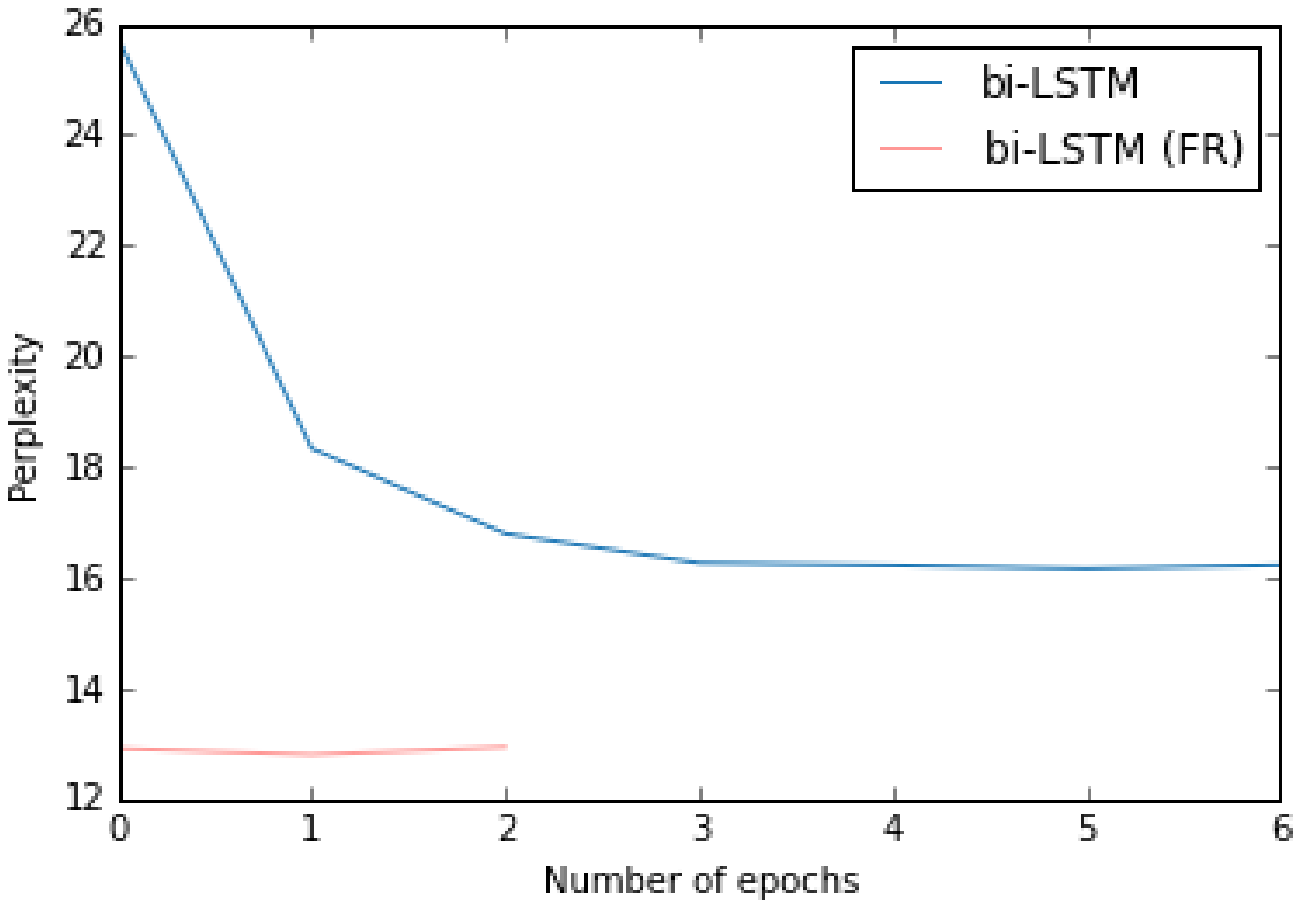}}
    \caption{Effect of cross-lingual pre-training for supersense tagging (left) and for low resource machine translation (right). Orange lines are the result from the model pre-trained with French.}
    \label{fig:effect}
\end{figure}

\subsection{Supersense Tagging}
Table \ref{tb:result-supersense} shows frequency weighted Precision, Recall F1 score\footnote{it can result in an F-score that is not between precision and recall.} on Semcor test set and Senseval3 all-words task. Our bidirectional LSTM model (bi-LSTM) outperformed the first sense heuristic baseline, the perceptron trained Hidden Markov Model proposed in~\citep{ciaramita2006broad}. And our new word-in-context pre-training model result in further improvements with all language pairs. The averaged score of 4 cross lingually pre-trained models, as in bi-LSTM (average), shows significant improvements over bi-LSTM. The model pre-trained with German achieved best result F1 84.1 on senseval3. Additionally, the baseline neural network models outperforms existing baselines even without cross lingual supervision.

\subsection{Lexical Translation in Low Resource Language}
Table \ref{tb:result-translation} shows results on machine translation in low resource language. We report the averaged BLEU score of 5 runs to avoid optimizer randomness~\cite{clark2011better}. The result show large improvement on perplexity and consistent improvement on BLEU in all language pairs. The average score of 4 cross lingually trained model improved perplexity by around 3 points and BLEU score by 0.3.

\subsection{Lexical Substitution}
Table \ref{tb:result-lexsub} shows results on lexical substitution task. Since our word-in-context representations are build only on Europerl parallel corpora, the baseline system is Skipgram word embedding trained on English side of EN-FR parallel corpora, which is the largest in the corpus. The Skipgram model which take most similar word as prediction is context in-sensitive baseline. Also we compared our results with various context sensitive models, which take  arithmetic mean (as in Add and BalAdd) and a geometrical mean (as in Mult and BalMult) of embeddings, proposed by~\citep{melamud2015simple}. They trained their baseline embeddings (as in Base) on a two billion word web corpus, ukWaC~\citep{ferraresi2008introducing}.

The model achieved best measures\footnote{evaluation was done by a script provided by the task organizer.} 10.63, best mode measure 18.90 with German supervision. And the second best result was obtained with Czech. As for comparison with~\cite{melamud2015simple}, we cannot compare score directly since we used different corpus and candidate generation. We should compare performance gain by taking into account context. Their best model (BalAdd) achieved 0.33 performance gain with context where our model achieved 2.9 performance gain on best evaluation.

\section{Discussion}

We proposed the model to predict lexical translation to build word-in-context representation. Table.~\ref{tb:example-lexsub} shows example of disambiguation with translation model in order of translation pribability. The model correctly disambiguate industrial plant (usine in French), and vegetable plant (plantes in French).  Figure \ref{fig:effect} shows the effect of pre-trained word-in-context representation for downstream tasks. Pre-trained model start from low perplexities at the first update and converged earlier, in two epochs, for low resource machine translation.

We investigated the effect of 4 linguistically diverse language. The results shows the benefit of cross-lingual pre-training in all languages, but overall the model trained with German have stable results and the model trained with Finnish tend to underperform others, especially on lexical substitution task where we do not have supervised fine-tuning process. This is probably because the large vocabulary of Finnish which is two times bigger than German.

\ignore{
As for model, it is also reasonable to  predict surrounding context in target language to disambiguate word senses. We pre-trained the model, which predict 6 surrounding context words of translation word with French and it underperformed our model on supersense tagging with F1 84.0 on Senseval3.}

\section{Related Work}
\paragraph{Word representation.} Distributed word representations were successfully applied to several downstream tasks such as chunking, parsing, sentiment analysis and paraphrase detection. Most of the tasks requires to use not only word representation but representation of phrases or documents. In the previous works, many architectures were proposed to learn and use word representation. In the sequence modeling problems such as BIO chunking, conditional random fields and recurrent neural networks are applied to represent a sequence of word representations~\citep{turian2010word,mesnil2013investigation}. For classification tasks such as document classification, sentiment analysis, paraphrase detection, summation of word embeddings~\citep{lauly2014learning}, convolutional neural networks~\citep{kalchbrenner2014convolutional} and recursive networks~\citep{socher2013recursive,cheng2015syntax} were proposed to represent compositionality function of words.

\paragraph{Learning semantics from parallel data.}  Previous works show methods to improve word or document level representation by incorporating multilingual context. \cite{faruqui2014improving} proposed canonical correlation analysis (CCA) based method to improve the quality of type level representation by projecting word representations of translation pairs (obtained by automatic word alignments) to be maximally correlated in common vector space. \cite{hermann2013multilingual} propose compositional vector space model (CVM) to build sentence representation. They represent a sentence as the sum of its word representations and they train word representation by constraining the representations of parallel sentences to be close. \ignore{Each model achieved good results on word-similarity task and  cross lingual document classification tasks respectively.} \cite{coulmance:2015} shows that predicting context in target language is an effective way to train word representation shared across languages. \cite{hill:2015} investigated the quality of word embedding learned by neural machine translation model and show its benefit on tasks that require modeling word similarity.

\paragraph{Compositional vector models.}
Most prior work on compositional vector models has looked primarily at the problem of computing representations of complete phrases rather than specifically words in context. Furthermore, one can learn reasonable generalizations from models that condition on and the generate text using an autoencoding objective~\citep{socher:2011}. \cite{dhillon:2012} make the intriguing proposition that left- and right- contexts can be used to supervise each other.

\ignore{
\noindent
\textbf{Supersense Tagging}: \cite{ciaramita2006broad} formulated supersense tagging as a sequence labeling problem. And they proposed Hidden Markov average perceptron algorithm over engineered spelling/morphological and contextual feature, such as previous label, word shape etc. Their model achieved an F-score 70.5 on Senseval-3 all words task which are used in this paper.
\cite{yuret2010noisy} proposed noisy channel model for unsupervised word sense disambiguation. They reported that their model achived F1-score 78 on Senseval-3, however this results are not compareble to the HMM model described above because they used different version of WordNet and evaluated only on noun examples in Senseval-3. For fair comparison, we can take the gap from first sense baseline and both models got 10\% improvements over the baseline.
}

\bibliography{iclr2016_conference}
\bibliographystyle{iclr2016_conference}

\end{document}